\documentclass[10pt,journal,compsoc]{IEEEtran}
\usepackage{cite}
\usepackage{amsmath,amssymb,amsfonts}
\usepackage{algorithm}
\usepackage{graphicx}
\usepackage{textcomp}
\usepackage{xcolor}
\usepackage{caption}
\usepackage{subcaption}
\usepackage{multirow}
\usepackage{multicol}
\usepackage{url}
\usepackage{booktabs}
\graphicspath{{./figures/}}
\usepackage{comment}
\usepackage{slashbox}

\def\A{{\bf A}}
\def\a{{\bf a}}

\def\e{{\bf e}}

\def\X{{\bf X}}

\def\x{{\bf x}}
\def\a{{\bf a}}

\def\0{{\bf 0}}
\def\1{{\bf 1}}

\def\tha{\mbox{\boldmath$\theta$\unboldmath}}

\newtheorem{theorem}{Theorem}

\newtheorem{definition}{Definition}

\newtheorem{cor}{Corollary}

\numberwithin{theorem}{section}
\numberwithin{lemma}{section}
\numberwithin{remark}{section}
\numberwithin{cor}{section}
\numberwithin{proposition}{section}

\usepackage[noend]{algpseudocode}

\algnewcommand\algorithmicinput{\textbf{INPUT: }}
\algnewcommand\Input{\item[\algorithmicinput]}
\algnewcommand\algorithmicoutput{\textbf{OUTPUT: }}
\algnewcommand\Output{\item[\algorithmicoutput]}

\renewcommand{\arraystretch}{1.3} 

\begin{document}

\title{Towards Private Learning on Decentralized Graphs with Local Differential Privacy}

 \author{Wanyu~Lin,~\IEEEmembership{Member,~IEEE,}
         Baochun~Li,~\IEEEmembership{Fellow,~IEEE,}
         and~Cong~Wang,~\IEEEmembership{Fellow,~IEEE}
 \IEEEcompsocitemizethanks{\IEEEcompsocthanksitem W.~Lin is with the Department of Computing, The Hong Kong Polytechnic University, Hong Kong, China. \protect\\ E-mail: wanylin@comp.polyu.edu.hk.

 \IEEEcompsocthanksitem B.~Li is with the Department
 of Electrical and Computer Engineering, University of Toronto, Ontario, M5S 1A1, Canada. \protect\\ Email: bli@ece.toronto.edu.
 \IEEEcompsocthanksitem C.~Wang is with the Department of Computer Science, City University of Hong Kong, Hong Kong, China. \protect\\ E-mail: congwang@cityu.edu.hk.

}
}


\IEEEtitleabstractindextext{%
\begin{abstract}
Many real-world networks are inherently decentralized. For example, in social networks, each user maintains a local view of a social graph, such as a list of friends and her profile. It is typical to collect these local views of social graphs and conduct graph learning tasks. However, learning over graphs can raise privacy concerns as these local views often contain sensitive information. 

In this paper, we seek to ensure private graph learning on a decentralized network graph. Towards this objective, we propose {\em Solitude}, a new privacy-preserving learning framework based on graph neural networks (GNNs), with formal privacy guarantees based on edge local differential privacy. The crux of {\em Solitude} is a set of new delicate mechanisms that can calibrate the introduced noise in the decentralized graph collected from the users. The principle behind the calibration is the intrinsic properties shared by many real-world graphs, such as sparsity. Unlike existing work on locally private GNNs, our new framework can simultaneously protect node feature privacy and edge privacy, and can seamlessly incorporate with any GNN with privacy-utility guarantees. Extensive experiments on benchmarking datasets show that {\em Solitude} can retain the generalization capability of the learned GNN while preserving the users' data privacy under given privacy budgets.
\end{abstract}

}

\maketitle

\IEEEdisplaynontitleabstractindextext
\IEEEpeerreviewmaketitle

\section{Introduction}
\label{sec:intro}

Many problems in scientific domains, ranging from computer networks~\cite{rusek2020routenet} and social networks to biomedicine and healthcare~\cite{zitnik2018modeling,zitnik2017predicting}, can be naturally cast as problems of property learning on graphs. Typically, these graph domains contain sensitive information. In computer networks, for example, the goal of botnets detection is to isolate the botnet nodes, where the problem can be formulated as binary node classification task on massive background Internet communication graphs~\cite{zhou2020auto}. However, the Internet Service Providers (ISPs) may be reluctant to share traffic observations (modeled as edges in the communication graphs~\cite{zhou2020auto}). Likewise, in social networks, users' contact lists, profile information, likes or comments, etc., should be kept be private, as most users are not willing to release their contact lists to strangers. Inevitably, the use of sensitive and private graph data requires principled and rigorous privacy guarantees.  
On the other hand, among various graph learning algorithms, graph neural networks (GNNs) have exhibited superior performance~\cite{zhou2020auto}, due to their efficiency and inductive learning capability~\cite{hamilton2017inductive}. Therefore, it is appealing to tailor the GNNs to perform graph learning tasks while still preserving user data privacy. A plausible approach is local differential privacy (LDP)~\cite{duchi2013local}, where each user locally obfuscates their share of data before sending them to a data curator (who may be malicious). As the data curator performs model learning over the obfuscated data, data privacy may be preserved under given privacy budgets. However, most existing LDP techniques for graphs mainly focused on graph statistics analysis while protecting the privacy of edge/link information. Typical graph statistics include subgraph counting~\cite{imola2021locally} (e.g., triangles and $k$-stars counting) and graph metric estimations~\cite{ye2020lf} (e.g., clustering coefficient, modularity, or centrality estimation), which are not designed for GNNs.

Learning over the obfuscated graphs with GNNs is quite challenging. In general, training deep learning models with strong differential privacy guarantees comes at a significant cost in utility~\cite{abadi2016deep,tramer2021differentially}. Specifically, in the context of graph learning, graph data usually contains {\em node feature} information and {\em graph structure} information. The combinatorial nature of graph structures makes the private learning problem more complex than other domains. These can be explained as follows. In GNNs, the node representations --- can be used for various downstream tasks --- are learned in a way that node information is aggregated and propagated via links through the message passing framework during training~\cite{hamilton2017inductive,kipf2016semi}. In node classification tasks or link prediction tasks, the training samples (nodes or links in the graph) of the learning model are interdependent. In contrast, the existing work on LDP for tabular data assumes that each user's data is independently and identically drawn from an underlying distribution~\cite{qin2016heavy,wang2017locally}, which is problematic in the context of graphs. 

\begin{figure}[t!]
   \centering
   \includegraphics[scale = 0.25]{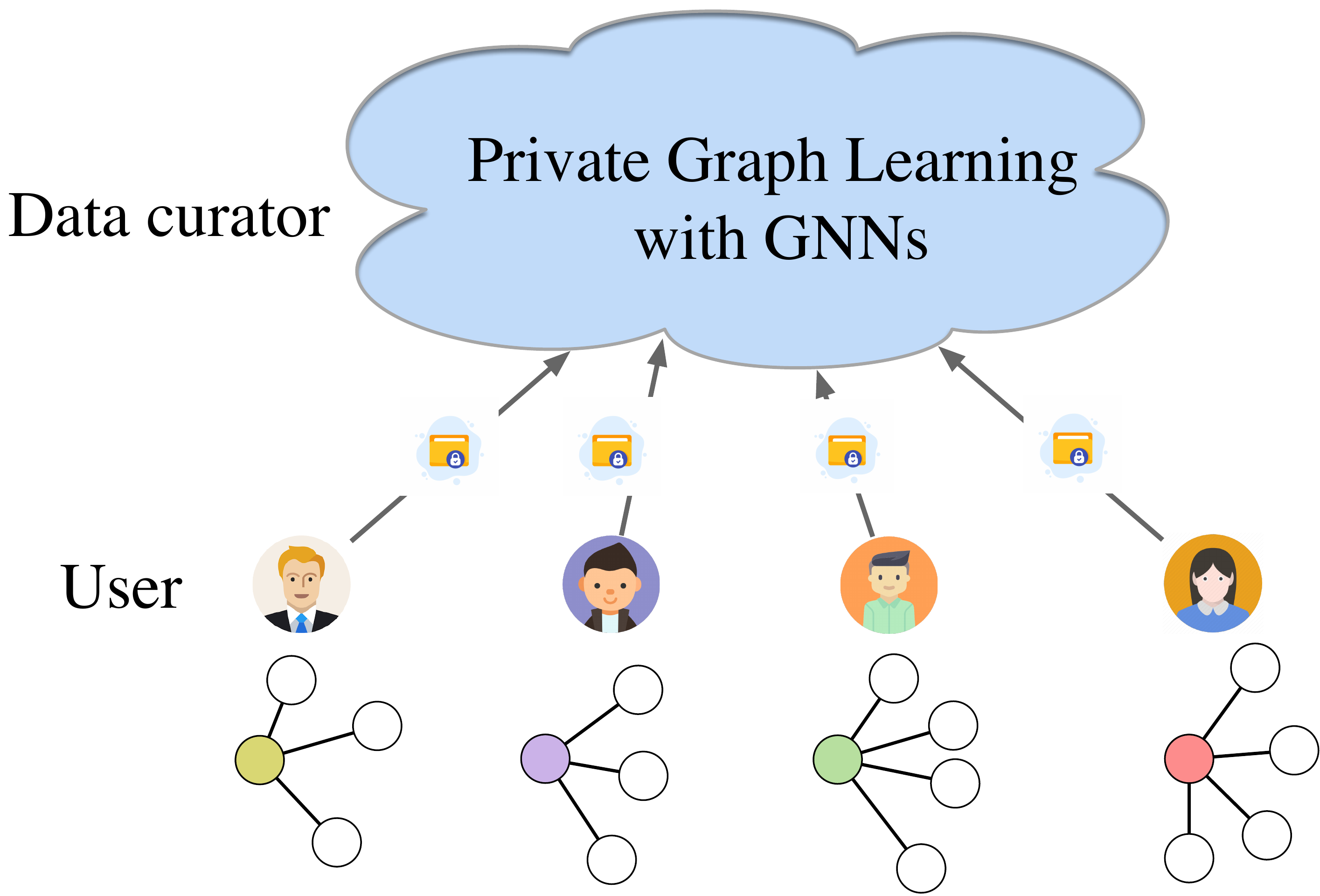}
  \caption{The illustration of private learning over decentralized graph network: 1) each user holds a data share including a local neighbour list and the profile information; 2) the data curator collects the obfuscated data and conducts learning over the collected noisy graphs.}
  \label{fig:model}
  \vspace{-10pt}
\end{figure}

While private graph learning with GNNs is still a nascent research topic, a recent proposal attempts to preserve the privacy of the node features~\cite{ccs2021}. However, this model raises several privacy and security issues as it assumes that the data curator holds the global graph structure. If the topological features contain sensitive information, this approach may incur information leakage as the data curator can directly access the global topology for the message passing process. Therefore, in this paper, we study the problem of differentially private graph learning with GNNs on a decentralized network graph, as shown in Fig.~\ref{fig:model}. In particular, the data curator cannot directly access the global structure of the graph. In other words, both the node feature and graph structure information should be protected against the data curator.

Privacy-preserving learning under the setting of decentralized network graphs has various applications, including but not limited to social network analysis and mobile computing. In principle, some social networks are inherently decentralized and distributed, such as Synereo~\cite{konforty2015synereo}. Though in some other social networks, e.g., Facebook, there exists a centralized party that holds the knowledge of the global network, that party may choose not to share it with a third-party (corresponding to the data curator) for analyzing, due to legal issues or other business concerns.

With the prevalence of decentralized network graphs, we propose a new framework consisting of a set of mechanisms, called {\em Solitude}, to tailor the GNNs for decentralized graph analysis under local differential privacy. Our framework has provable privacy guarantees based on local differential privacy. Specifically, we leverage the notion of edge local differential privacy proposed in~\cite{qin2017generating}. To protect the privacy of neighbor lists, each user applies Warner's randomized response mechanism~\cite{dwork2014algorithmic} to obfuscate their neighbor lists before sending them to the data curator. For protecting node/user\footnote{In many applications, such as social networks, each node represents a user; we use ``node'' and ``user'' interchangeably.} feature privacy, we further incorporate a multi-bit mechanism for multi-dimensional feature perturbation~\cite{ding2017collecting,ccs2021}, which can ensure that the high-dimensional feature vector of every user can be protected. However, learning over obfuscated graphs introduces challenges, as it could significantly degrade the generalization capability of the GNN. Precisely, the learned GNN may overfit the noisy graphs and generalize poorly to unseen nodes.  

To this end, we propose new mechanisms for graph structure calibration and feature vector calibration, respectively. Essentially, obfuscating graph structure via a randomized response mechanism introduces edge deletion, addition, and rewiring. Though the obfuscating process satisfies our privacy goal, we theoretically and empirically analyze that this process tends to return a much denser graph than the original one, violating the sparsity property --- an inherent property exhibited in many real-world graphs~\cite{zhu2019robust}. Therefore, we propose encouraging the sparseness of the graph structure during training as a calibration step to reduce the effect of the noise. In addition, a node might be linked to nodes with task-specific ``noise,'' leading to the aggregation of non-smooth features in GNNs. Inspired by the assumption of {\em feature smoothness} in GNNs, we leverage a feature smoothing component before training to reduce the effect of the feature noise. To further boost the prediction accuracy, a label smoothness component is adopted, which is inspired by the principles of label propagation algorithms~\cite{bengio200611,zhou2004learning}.

In a nutshell, our original contributions are listed as follows. We propose a new privacy-preserving learning framework for decentralized network graphs based on graph neural networks. It consists of a set of mechanisms that can provide local differential privacy for every user yet maintaining the generalization capability of the learned GNN. We formally analyze that our mechanisms preserve local differential privacy for every user, particularly on the notion of edge differential privacy. Towards our goal, we theoretically and empirically analyze the overfitting problem while learning over the noisy graphs. Different from the literature on locally private GNNs, {\em Solitude} can preserve edge privacy and node feature privacy for every user simultaneously. Our extensive array of experiments on benchmarking datasets demonstrated that {\em Solitude} can significantly improve the privacy-utility guarantees on canonical graph learning benchmarks. We also empirically show that our framework can seamlessly integrate with any GNN architectures, such as GCN~\cite{kipf2016semi} and GraphSage~\cite{hamilton2017inductive}, with privacy-utility guarantees.


\section{Problem Setup}

\subsection{Notations and Problem Definition} 
 \begin{table}[t]
\caption{Notations}
\begin{center}
\setlength{\tabcolsep}{9pt} 
\renewcommand{\arraystretch}{1.17} 
\begin{tabular}{ c| c}
\hline
Notation & Descriptions \\ 

\hline
$\a_i$ & the adjacency list of user i\\
\hline
$\tilde{\a}_i$ & randomized adjacency list of user i\\
\hline
$\x_i$ & the feature vector of user i\\
\hline
$\hat{\x}_i$ & the encoded feature vector of user i\\
\hline
$\tilde{\x}_i$ & the rectified feature vector of user i\\
\hline
$\mathcal{M}_{a}$ &the randomized mechanism for adjacency lists\\
\hline
$\mathcal{M}_{x}$ &the randomized mechanism for node features\\
\hline
$\epsilon_x,\, \epsilon_a$ &privacy budgets for features and adjacency lists \\
\hline
$|\mathcal{N}(\cdot)|$& the node degree of user i\\
\hline
$\A^c$& the adjacency matrix after calibration\\
\hline
$\X^c$& the node feature matrix after calibration\\
\hline
$||\cdot||^2_{\mathbf{F}}$& the Frobenius norm of a matrix\\
\hline
$||\cdot||_1$ &$l_1$ norm operator\\
\hline
$\tha$ &the model parameters\\
\hline
$\lambda_1,\,\lambda_2$& the regularization coefficients\\
\hline
\end{tabular}
\end{center}
\vspace{-10pt}
\label{tab:notations}
\end{table}

{\bf Notations.} We consider a network graph, denoted as $\mathcal{G}=(\mathcal{V},\, \A,\,\X)$. Specifically, we consider the decentralized setting, where the entire graph is decentralized over users/nodes $\mathcal{V}=\{v_1,\cdots,v_{|\mathcal{V}|}\}$. Precisely, each user $v_i$ holds locally a neighbor list depicted as $\a_i$, which can be modeled as an $|\mathcal{V}|$-binary vector, and a node attribute/feature vector $\x_i\in\mathbb{R}^{|D|}$, where $|D|$ is the dimension of the user/node feature vector. The corresponding adjacency matrix of the entire graph can be represented as $\A=\{\a_1,\cdots,\a_{|\mathcal{V}|}\}$, where $\A\in\mathbb{R}^{|\mathcal{V}|\times|\mathcal{V}|}$, and the node attribute matrix $\X=\{\x_1,\cdots,\x_{|\mathcal{V}|}\}$, where $\X\in \mathbb{R}^{|\mathcal{V}|\times |D|}$.  In the context of social network graphs, for example, each user locally holds their friend list and the profile information (e.g., WeChat~\cite{wechat}). Without loss of generality, we assume $\mathcal{G}$ is a directed graph; this reflects the real-world application domains, e.g, the follower-followee relationships.

Consistent with prior work~\cite{ccs2021}, we focus on the node classification task with graph neural networks. Specifically, we follow the standard node classification setting, which is commonly employed in various literature~\cite{kipf2016semi,hamilton2017inductive}. Given a set of labeled nodes $\mathcal{V}_l\subset \mathcal{V}$, with class labels from $\mathcal{Y}=\{y_1, y_2, y_3,\cdots, y_K\}$ and a set of unlabeled nodes $\mathcal{V}_u \subset \mathcal{V}/ \mathcal{V}_l$, the goal of node classification is to map each node $v\in\mathcal{V}$ to one class in $\mathcal{Y}$.

{\bf Problem Definition.} We assume that the data curator is an untrusted party, and it can access the set/index of nodes/users $\mathcal{V}=\{v_1,\cdots,v_{|\mathcal{V}|}\}$ and the node labels of the training set $\mathcal{V}_l$. However, the data curator could not directly access the feature matrices $\X$ and $\A$, which are decentralized among the users and private to the users. The data curator is allowed to collect information from the users, and performs node classification with GNNs over the noisy graphs. 

Therefore, our ultimate goal is to obtain a set of mechanisms that 1) the data curator can collect data portions from each user while preserving the user data privacy; 2) the data curator can train a GNN for node classification over the collected noisy graph with the best possible generalization capability. Without loss of generality, model generalization capability is measured by the prediction accuracy on the held-out test set that has not been seen during training. Note that, different from the state-of-the-art on locally private GNN~\cite{ccs2021} --- LPGNN, we consider a more advanced setting that the global topology of the graph is not accessible to the data curator. Concretely, user $v_i$'s private data includes the neighbor list, or called {\em adjacency list}, $\a_i$ and the node attribute/feature vector $\x_i$. In what follows, we first briefly introduce some necessary background on graph learning with GNNs and the notion of local differential privacy on graphs to facilitate a better understanding of our solution. The mathematical notations used in this paper are summarized in Table~\ref{tab:notations}.

\subsection{Message Passing Graph Neural Networks}

Graph neural networks (GNNs) are tailored to learn and model information structured as graph data. It is a family of graph message passing architectures that incorporate graph structure and node feature vectors to learn a dense representation of a node or the entire graph. In principle, GNNs share a neighborhood aggregation strategy, where the node representations are refined via iteratively aggregating the representations from its neighboring nodes in the graph. Representative GNNs are graph convolutional networks, which use mean pooling for aggregation~\cite{kipf2016semi}, and GraphSage that aggregates the node features via mean/max/LSTM pooling~\cite{hamilton2017inductive}. 

Taking GCNs as an example, the basic operator for the neighborhood information aggregation is the element-wise $\mathbf{mean}$. After $L$ iterations of aggregation, a node’s representation can capture the structural information within its $L$-hop graph neighborhood, which can be formulated as:
 \begin{equation}
  h^{l+1}_v\,\leftarrow\,\sigma\,\left(W^l\,\cdot\,\mathbf{mean}\left(\{h_v^{l}\}\cup\,\{h_u^{l},\,\forall\,u\in\,\mathcal{N}(v)\}\right)\right)
 \end{equation}
where $W^l$ is a trainable matrix for layer $l$, $\sigma$ denotes a nonlinear activation function. Note that, except for $\mathbf{mean}$ operator, there are many other operators for neighborhood information aggregation, such as $\mathbf{max}$ pooling. Due to their superior performance, these operators have been the core of many graph neural networks. For more details on other GNN variations, we refer the interested readers to the existing survey~\cite{wu2020comprehensive}.

\subsection{Local Differential Privacy}
\label{subsec:ldp}

Local differential privacy (LDP) has emerged as the {\em de facto} solution for collecting private data and performing statistical queries, such as mean, counting, etc. In principle, it is a privacy metric to protect the sensitive information of individuals from the data curator~\cite{raskhodnikova2008can,duchi2013local}. In the setting of LDP, each user does not trust the data curator. Before sending her share of data to the data curator, each user locally perturbs the data portion with a differentially private mechanism. The perturbed data is not meaningful individually but can be used for data analytics when aggregated. 

In the context of graph data, a differentially private mechanism can be designed for edge differential privacy~\cite{blocki2012johnson}, or node differential privacy~\cite{day2016publishing}. In essence, edge differential privacy ensures that a randomized mechanism does not reveal the addition or deletion of an edge in the neighbor list of an individual. In contrast, a randomized mechanism for node differential privacy hides the deletion or addition of a node along with its link list. Consistent with prior works~\cite{qin2017generating,ye2020lf}, we adopt the notion of edge local differential privacy (LDP) based on a user's neighbor list. 

Formally, let $\a_i=(a_{i,1},\cdots,a_{i,|\mathcal{V}|})\in\{0,1\}^{|\mathcal{V}|}$ be the neighbor list of the user $v_i$, where $\a_i$ is the $i$-th row of the adjacency matrix $\A$ of the global network graph $\mathcal{G}$. Stated differently, the adjacency matrix of $\mathcal{G}$ can be represented as $\A=\{\a_1, \a_2,\cdots,\a_{|\mathcal{V}|}\}$. Then edge LDP can be defined as follows~\cite{qin2017generating,imola2021locally}:

\begin{definition}[\textbf{Edge local differential privacy}]\label{def:eLDP}
A randomized mechanism $\mathcal{M}$ satisfies $\epsilon$-edge local differential privacy ($\epsilon$-edge LDP) if and only if for any two neighbor lists $\a$ and $\tilde{\a}$, such that $\a$ and $\tilde{\a}$ only differ in one bit, and any $s\subseteq \mathbf{range}(\mathcal{M})$, we have 
\begin{equation}\frac{\mathbf{Pr}[\mathcal{M}(\a)=s]}{\mathbf{Pr}[\mathcal{M}(\tilde{\a})=s] } \leq e^{\epsilon},
\end{equation} 
where $\epsilon$ is the privacy budget.
\end{definition}

Note that, $\epsilon$-edge LDP in Definition~\ref{def:eLDP} protects one single bit in a neighbor list with privacy budget $\epsilon$. By adopting the notion of group privacy~\cite{dwork2014algorithmic}, $\epsilon$-edge LDP can be used to protect $k\in\mathbb{N}$ edges. Concretely, if $\mathcal{M}$ provides $\epsilon$-edge LDP, then for any two neighbor lists $\a$ and $\tilde{\a}$ that differ in $k$ edges and any $s\subseteq \mathbf{range}(\mathcal{M})$, we have 
\begin{equation}\frac{\mathbf{Pr}[\mathcal{M}(\a)=s]}{\mathbf{Pr}[\mathcal{M}(\tilde{\a})=s] } \leq e^{k\epsilon},
\end{equation}
where $k$ edges are protected with privacy budget $k\epsilon$.

{\em Properties.} LDP satisfies composition property and transformation invariance~\cite{day2016publishing}. Specifically, composition property enables the modular design of mechanisms: if all the components of a mechanism are differentially private, so is their composition. The transformation invariance demonstrates that performing post-processing on the output of an algorithm that satisfies LDP does not affect the privacy guarantee.

\section{{\em Solitude}: our proposed framework}
\label{sec:protocol}
\begin{figure*}[t!]
   \centering
   \includegraphics[scale = 0.21]{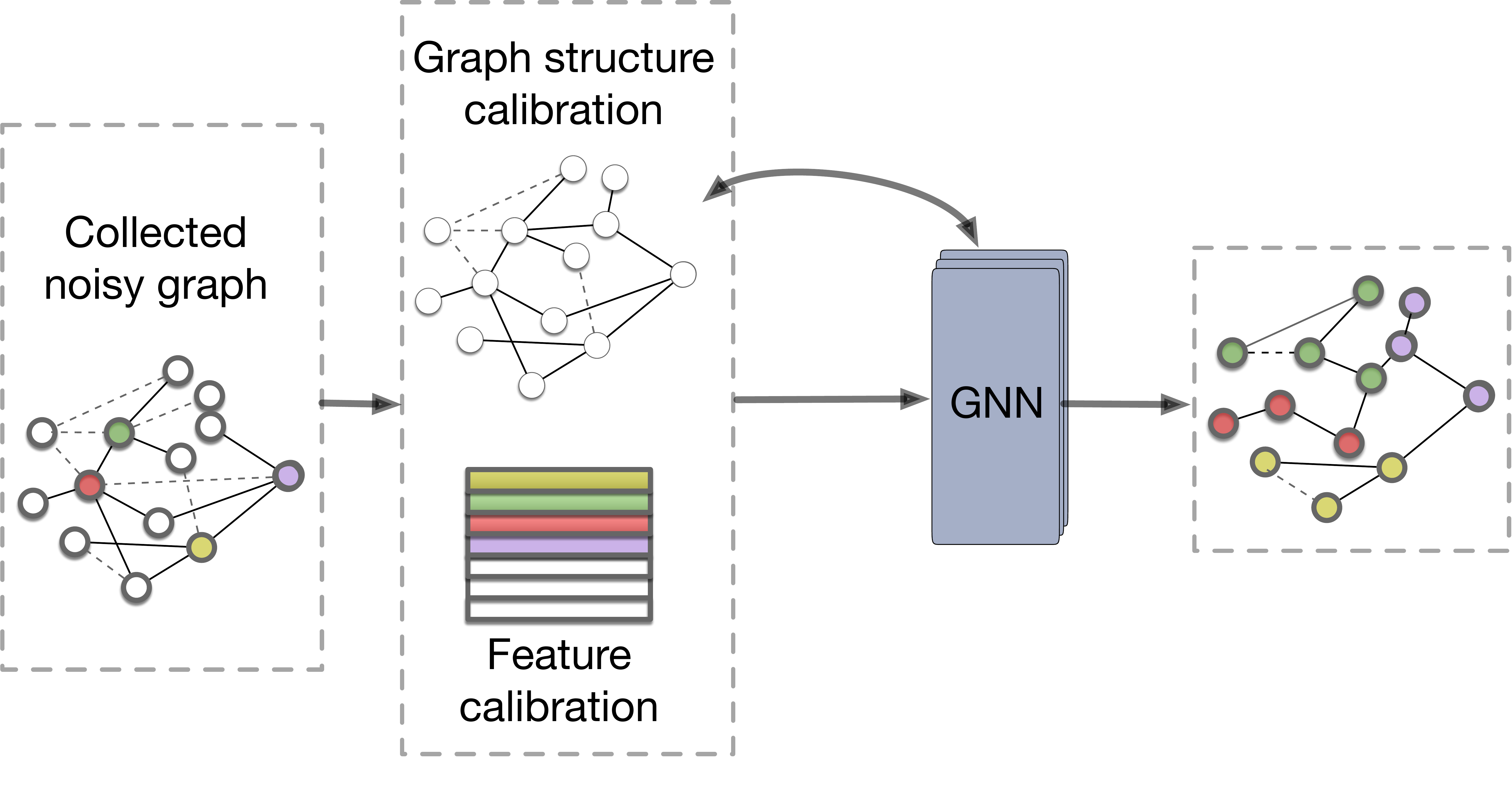}
      \vspace{-15pt}
   \caption{Illustration of the private GNN training at the data curator side. It consists of two components: 1) a feature denoising component to reduce the effect of non-smooth aggregation; it is carried out before the training process; 2) a graph structure denoising component to enforce the graph structure's sparseness; this process is jointly optimized with the model training.}
  \label{fig:framework}
\vspace{-15pt}
\end{figure*}

This section describes the main components of our proposed framework, called {\em Solitude}, toward differentially private training of GNN over private graph data, including the node feature vectors and the neighbor lists that are decentralized across all users. Specifically, there are two stages in our framework. In the first stage, the data curator sends a query to each user $v_i$ once, and then each user $v_i$ independently sends an answer -- the obfuscated data share, including the obfuscated version of node feature vector $\tilde{\x}_i$ and the obfuscated adjacency list $\tilde{\a}_i$ (shown in Fig.~\ref{fig:model}). In the second stage, the data curator performs training of the GNN over the noisy graph composed of the obfuscated data shares from all users. The second stage of our {\em Solitude} is illustrated in Fig.~\ref{fig:framework}. 

In what follows, we first introduce the technical details of the randomized mechanisms used for preserving data privacy satisfying local differential privacy. Then we theoretically and empirically analyze the overfitting issue caused by the randomized mechanisms. We show how to calibrate the introduced noise such that the generalization capability of the trained GNN can be retained.

\subsection{Obfuscating Local Data under Local Differential Privacy}

In the setting of the decentralized network graph, each user $v_i$ holds a data portion of the entire graph, including their feature vector $\x_i$ and the adjacency list $\a_i$, both of which are private to the user. In the subsequent paragraphs, we describe the used mechanisms for protecting the privacy of the feature vector and the adjacency list, respectively.  

{\bf Randomized Adjacency List.} Intrinsically, an adjacency list is a binary bit vector, denoted as $\a_i=\{a_{i,1}, \cdots, a_{i,|\mathcal{V}|}\}$, in which $a_{i,j}=1$ indicates the link between $v_i$ and $v_j$. Without loss of generality, we leverage of a common methodology, called {\em randomized response}~\cite{qin2017generating,dwork2014algorithmic}, to impel local differential privacy. Specifically, each user flips each bit of her adjacency list with a probability $p$ constrained by a given privacy budget. More formally, given a privacy budget $\epsilon_a$, the randomized adjacency list, denoted as $\tilde{\a}_i=\{\tilde{a}_{i,1}, \cdots, \tilde{a}_{i,|\mathcal{V}|}\}$, is obtained as follows:

\begin{equation}\label{eq:ral}
  \tilde{a}_{i,j} = \left\{
\begin{array}{ll}
      a_{i,j}, & q = \frac{\e^{\epsilon_a}}{1+\e^{\epsilon_a}} \\
      1-a_{i,j},& p =\frac{1}{1+\e^{\epsilon_a}} \\
\end{array} 
\right.,
\end{equation}
where $q=1-p$ is the probability of retaining a particular bit in the adjacency list.

\begin{theorem}\label{theorem:edge-ldp}
The randomized adjacency list mechanism $\mathcal{M}_{a}$ satisfies $\epsilon_a$-edge local differential privacy.
\end{theorem}

{\em Proof.} Let us consider the case that $\a_i$ and $\tilde{\a}_i$ only differ in one bit. Concretely, we assume that $a_{i,j}\neq\tilde{a}_{i,j}$, and given any output $s=(s_1,\cdots,s_n)$ from $\mathcal{M}$, we have 
\begin{align}\frac{\mathbf{Pr}[\mathcal{M}_a(\a_i)=s]}{\mathbf{Pr}[\mathcal{M}_a(\tilde{\a}_i)=s] }=\frac{\mathbf{pr}[a_{i,1}\rightarrow s_1)]\cdots \mathbf{pr}[a_{i,n}\rightarrow s_n)]}{\mathbf{pr}[\tilde{a}_{i,1}\rightarrow s_1)]\cdots \mathbf{pr}[\tilde{a}_{i,n}\rightarrow s_n)]} \\
=\frac{\mathbf{pr}[a_{i,j}\rightarrow s_j)]}{\mathbf{pr}[\tilde{a}_{i,j}\rightarrow s_j)]}=\frac{q}{p}\leq e^{\epsilon_a}.
\end{align} 

\begin{figure}[h]
   \centering
   \includegraphics[scale = 0.5]{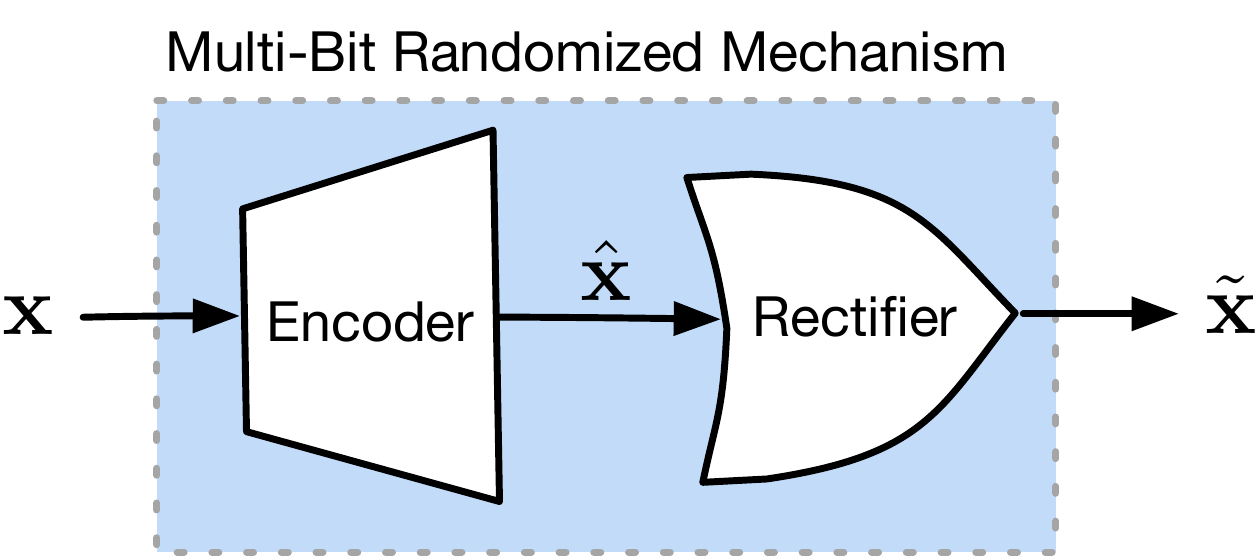}
   \caption{Illustration of multi-bit mechanism. The encoder encodes the multi-dimensional features $\x_i\in\mathcal{R}^{|D|}$ into $\hat{\x}_i\in\{-1,0,1\}^{|D|}$, with a probability defined by the privacy budget. The rectifier is to calibrate the encoded vector $\hat{\x}_i$ to ensure the outcome $\tilde{\x}_i$ is statistically unbiased.}
  \label{fig:bit-vector}
  \vspace{-10pt}
\end{figure}

{\bf Randomized Feature Vector.} If the privacy of the node feature is also the concern, we further leverage a multi-bit mechanism to protect its privacy for every user. The randomized mechanism for feature vector in our framework follows the similar outline as the mechanism used in\cite{ccs2021,ding2017collecting}. Specifically, the multi-bit mechanism has two components: an encoder and a rectifier, as shown in Fig.~\ref{fig:bit-vector}. To randomize a feature vector $\x_i\in\mathcal{R}^{|D|}$, where each element $x_{i,j}$ falls into the range $[x_{\mathbf{min}}, x_{\mathbf{max}}]$, the encoder first uniformly samples $m$ features out of the $D$ dimensions. Each of the selected features is encoded into $-1$ or $1$, with a probability formulated as
\begin{equation}\label{eq:encoder}
\frac{1}{\e^{{\epsilon_x}/m}+1}+\frac{x_{i,j}-x_{\mathbf{min}}}{x_{\mathbf{max}}-x_{\mathbf{min}}}\cdot\frac{\e^{{{\epsilon_x}/m}}-1}{\e^{{{\epsilon_x}/m}}+1}.
\end{equation} 
Correspondingly, the rest of the $d-m$ features are mapped to $0$s. The rectifier is to calibrate the encoded vector $\hat{\x}$ to ensure the outcome of the randomized mechanism $\tilde{\x}$ is statistically unbiased. Formally, the rectifier is instantiated as
\begin{align}\label{eq:rectifier}
\mathbf{Rec}(\hat{x}_{i,j})=\frac{|D|\cdot(x_{\mathbf{max}}-x_{\mathbf{min}})}{2m}\cdot\frac{\e^{{{\epsilon_x}/m}}+1}{\e^{{{\epsilon_x}/m}}-1}\cdot{\hat{x}_{i,j}}\\
+\frac{x_{\mathbf{max}}+x_{\mathbf{min}}}{2}.
\end{align}

\begin{theorem}\label{theorem:feature-ldp}
The randomized mechanism for node feature vector $\mathcal{M}_{x}$ preserves $\epsilon_x$ differential privacy of every user.
\end{theorem}
Due to the page limitation, we refer to the detailed proof of Theorem~\ref{theorem:feature-ldp} to~\cite{ding2017collecting,ccs2021}. Note that the multi-bit mechanism for feature vector is different from the notion of node local differential privacy (node-LDP). Specifically, node-LDP hides the deletion or addition of a node along with its neighbor list; it is out of the scope in this paper. With the composition property of local differential privacy as described in Sec.~\ref{subsec:ldp}, we arrive at the following corollary:
\begin{cor}\label{corollary:ldp}
 The randomized mechanisms ($\mathcal{M}_{a}$ and $\mathcal{M}_{x}$) together satisfy $\epsilon_a+\epsilon_x$ differential privacy of every user.
\end{cor}

\subsection{Private GNN Training with Calibration}

Now with the collected graph data from all users, the data curator can reconstruct the global graph, and it is ready to perform graph analytics with message passing GNNs; the task is instantiated with node classification in this work. Yet it can be generalized to other graph learning tasks, such as link prediction, clustering coefficient prediction~\cite{you2021identity}, as these tasks all share the same graph message passing architecture. Unfortunately, the collected graph is noisy; aggregating and propagating the noisy information leads to over-fitting which degenerates the generalization ability.

{\bf Why does the performance degrade?} Although the randomized collection of the adjacency lists and feature vectors satisfies our privacy goal, the reconstructed graph does not reflect the original decentralized social graph well. From the perspective of the graph structure, it tends to return a much denser graph than the original one. For example, a citation network Cora~\cite{mccallum2000automating} has $|\mathcal{V}|=2708$ nodes, and the average node degree is $3.89$. Mathematically, the randomized mechanism $\mathcal{M}_{a}$ introduces $p\times |\mathcal{V}|^2=6681$ edge flipping with $\epsilon_a=7$ in expectation. We empirically analyze the average node degree of the reconstructed graph in the setting of $\epsilon_a=7$; it is $6.35$ by averaging the results of $5$ repeats. Concretely, it introduces $6634$ edges that may be task-irrelevant, averaged over $5$ repeats. A node might be linked to nodes with task-specific ``noisy'' edges due to the randomized flipping. Aggregating messages from these nodes would compromise the quality of the node embedding and lead to undesirable predictions in the downstream tasks.  

To visualize the performance degradation, we evaluated the task of document classification with different values of privacy budgets on Cora~\cite{mccallum2000automating} and CiteSeer~\cite{giles1998citeseer} respectively. As observations are similar in other values of $\epsilon_x$, we present the results of $\epsilon_x=1$ while changing the value of $\epsilon_a$ in Table~\ref{tab:cora}. We observed that under the randomized mechanisms, the performance of GNNs is non-significant. In what follows, we are interested to exploit the intrinsic properties of the data to boost the classification accuracy with further gains when the graph is collected satisfying given privacy goals. 

\begin{table}[!ht]
\caption{Classification accuracy ($\%$) on Cora and CiteSeer, with $\epsilon_x=1$. The performance are insignificant in various edge privacy budgets.}
\vspace{-10pt}
\begin{center}
\begin{small}
\begin{sc}
\setlength{\tabcolsep}{0.17em}
\begin{tabular}{ c |c |c |c| c|c}
\toprule
 $\epsilon_{a}$& 7.0 & 7.3 & 7.5& 7.7  & 7.9 \\ 
\midrule
 Cora&$50.9\pm3.8$ & $54.9\pm3.7$ & $57.2\pm2.5$  & $61.4\pm2.6$& $63.7\pm2.4$ \\
 CiteSeer&$35.0\pm1.8$&$39.1\pm1.3$&$42.3\pm1.5$&$44.5\pm1.2$&$45.7\pm1.8$\\
\bottomrule
\end{tabular}
\end{sc}
\end{small} 
\end{center}
\vspace{-10pt}
\label{tab:cora}
\end{table}

{\bf Graph Structure Denoising.} The randomized flipping introduces ``noisy'' edges that can degrade the generalization performance of the GNNs. These edges tend to connect nodes within different communities or with different labels, and they should be pruned for better learning performance. According to the analysis mentioned above, the randomized flipping increases the density of the graph. Therefore, we propose to calibrate the collected noisy graph by encouraging the sparseness of the graph structure. In particular, we calibrate the graph structure by minimizing the $l_1$ norm of the calibrated adjacency matrix, denoted as $||\A^c||_1$. 
Therefore, the graph structure calibration process can be formulated as an optimization problem, shown in Eq.~\ref{eq:a_calibration}.

\begin{equation}\label{eq:a_calibration}
\min_{\A^c} ||\tilde{\A}-\A^c||^2_{\mathbf{F}}+\lambda||\A^c||_1,
\end{equation}
where $\A^c$ represented the adjacency matrix after calibration, and $\lambda$ control the associated calibration level. The first term is to ensure the calibrated matrix to be close to the collected graph topology. Concretely, the distance of the calibrated matrix and the collected matrix is measured by the Frobenius norm. The Frobenius norm of a matrix $\A$ is defined by $||\A||^2_{\mathbf{F}}=\Sigma a_{i,j}^2$. We are aware of the drawbacks of reusing notations. $\A$ in the definition of the Frobenius norm represents any matrix for simplicity.

{\bf Node Feature Vector Denoising.} In GNNs, it computes the new representation of a node by aggregating and propagating information from its neighbors~\cite{hamilton2017inductive}. The learned representations of connected nodes tend to be similar. Stated differently, for message passing graph neural network to work, a certain assumption, called the {\em smoothness assumption}, has to hold. This reflects the real-world phenomenon on graphs from various domains. For example, two connected users in a social graph are likely to share similar features, fulfilling the property of {\em feature smoothness}. However, in the noisy graphs, the GNNs may result in degenerated node representations due to the non-smooth features aggregating from the neighbors, leading to performance degradation of the learned GNN.

To address the above issue, we leverage a feature smoothing component that applies a mean aggregator~\cite{ccs2021} to enhance the node features -- denoising via mean aggregation of the node features from the neighbors. Specifically, instead of using the rectified features $\tilde{\X}$, we refine the features of each node by averaging the feature vector from their neighbors within $l$-hop. Formally, the process of feature smoothing within $1$-hop can be formulated as:
\begin{equation}
\label{eq:denoise_feat}
\x_i^c =\sum_{v_j\in\mathcal{N}(v_i)}^{} \frac{\tilde{\x}_j}{|\mathcal{N}(v_i)||\mathcal{N}(v_j)|}.
\end{equation}
Our feature smoothing component is executed with $l_x$ times, which is data-driven and needs to be tuned to avoid over-smoothing problem. 

These denoising processes, including graph-structure denoising and feature vector denoising, are designed as calibration steps for better learning of the GNNs. According to the transformation invariance of LDP~\cite{day2016publishing}, these processes would not affect the privacy guarantee. Stated differently, the output of these processes is still noisy and does not reflect the private data portion of each user. Nevertheless, the feature smoothing operation is designed to reduce the noise effect that may degrade the generalization capability of the GNN.   

{\bf Model Training.} To evaluate the effectiveness of {\em Solitude} for privacy-preserving graph learning tasks, we instantiate the graph learning task with node classification. We denote the target classifier as $f(\tilde{\x})=\mathbf{arg}\,\max_{y}p(y|\tilde{\x})$, where $p(y|\tilde{\x})=g(\A^c; \X^c, \tha)$. More specifically, $g$ represents the GNN model and $\tha$ are the learned model parameters given the obfuscated graph $\tilde{\mathcal{G}}=(\tilde{\A},\,\tilde{\X})$ and the denoising mechanisms. Note that the generalization capability of the learned model is measured by the prediction accuracy over an obfuscated/noisy test set that is not seen during training. To further boost the prediction accuracy, we further incorporate a label smoothing component, which is inspired by the principles of label propagation algorithms~\cite{bengio200611,zhou2004learning} --- node labels are propagated and aggregated along edges in the graph. Formally, the label smoothing component within $1$-hop is formulated as:

\begin{equation}
\label{eq:denoise_label}
p(y^c|\tilde{\x}) =\sum_{v_j\in\mathcal{N}(v_i)}^{} \frac{p(y|\tilde{\x})}{|\mathcal{N}(v_i)||\mathcal{N}(v_j)|}.
\end{equation}
The label smoothing component can be carried out with $l_y$ times, similar to the feature smoothing component. In Sec.~\ref{sec:experiments}, we will show that for node classification tasks, improving privacy-utility guarantees needs more labeled samples for training.

In general, the model parameters are obtained by minimizing the cross-entropy error over all labeled samples:
\begin{equation}
\label{eq:gcnloss}
\mathcal{L}_{\mathbf{GNN}}(\A^c, \X^c, \tha) = -\sum_{v_i\in\mathcal{V}_l} \sum_{k=1}^{K} y_{ik}\ln p(y^c_k|\tilde{\x}),
\end{equation}
where $\mathcal{V}_l$ is the set of node indices that have labels, and K is the number of classes/labels. 

Intuitively, the denoising processes can be regarded as data preprocessing for model training. In other words, at the data curator side, {\em Solitude} can first proceed with the mechanism for feature vector denoising and then optimize Eq.~\ref{eq:a_calibration} to obtain a ``denoised'' graph structure. After that, the ``denoised'' graph is treated as the input of the GNN model. By solving Eq.~\ref{eq:gcnloss}, the model parameters $\tha$ can be obtained. Note that our ultimate goal is to obtain a GNN model with optimum generalization capability measured by the prediction accuracy on the test set. From this perspective, we can treat the process of graph structure denoising as a form of regularization, which is aligned with the regularization technique that prevents neural networks from overfitting~\cite{kukavcka2017regularization}. Accordingly, the loss function for model training can be reformulated as:

\begin{equation}
\label{eq:finalloss}
\min_{\A^c,\,\tha} \mathcal{L}_{\mathbf{GNN}}(\A^c, \X^c, \tha) +\lambda_1||\tilde{\A}-\A^c||^2_{\mathbf{F}}+\lambda_2||\A^c||_1,
\end{equation}
where $\lambda_1$ and $\lambda_2$ control the regularization ratio. To solve Eq.~\ref{eq:finalloss}, an alternating optimization scheme with Adam is used to iteratively update $\tha$ and $\A^c$. As there is no parameter learning during feature denoising, the feature smoothing component will proceed before model training.

\begin{table*}[!t]

\caption{Classification accuracy ($\%$) with $\epsilon_x=1$, on various values of $\epsilon_a$. }
\begin{center}
\begin{small}\small
\begin{sc}
\renewcommand{\arraystretch}{1.3} 

\begin{tabular}{ c |c| c | | c| c |c| c |c| c}
\toprule
Dataset&&$\epsilon_a$&$7.0$ &$7.3$& $7.5$&$7.7$&$7.9$& $8.0$\\ 
\midrule
\multirow{6}{*}{Cora} 
&\multirow{3}{*}{GraphSage}
&Base &$50.9\pm3.8$&$54.9\pm3.7$&$57.2\pm2.5$&$61.4\pm2.6$&$63.7\pm2.4$&$63.2\pm3.2$\\
&&LPGNN &$62.5\pm1.6$&$69.2\pm1.6$&$72.9\pm2.3$&$74.7\pm0.9$&$75.3\pm1.2$&$76.4\pm0.8$\\
&&{\em Solitude} &$\mathbf{66.4}\pm1.7$&$\mathbf{72.4}\pm1.2$&$\mathbf{75.8}\pm0.9$&$\mathbf{76.7}\pm1.1$&$\mathbf{77.9}\pm0.6$&$\mathbf{79.0}\pm0.6$\\
\cline{2-9}
&\multirow{3}{*}{GCN}
&Base &$60.8\pm2.0$&$63.1\pm3.4$&$64.8\pm3.7$&$66.9\pm4.0$&$68.1\pm4.0$&$68.6\pm3.8$\\
&&LPGNN &$67.9\pm1.2$&$69.4\pm1.4$&$73.7\pm1.8$&$74.4\pm0.5$&$75.7\pm1.0$&$76.3\pm0.9$\\
&&{\em Solitude} &$\mathbf{68.4}\pm2.3$&$\mathbf{72.6}\pm1.4$&$\mathbf{75.2}\pm1.2$&$\mathbf{76.1}\pm0.5$&$\mathbf{77.3}\pm0.9$&$\mathbf{77.8}\pm0.5$\\
\hline
\multirow{6}{*}{CiteSeer} 
&\multirow{3}{*}{GraphSage}
&Base &$35.0\pm1.8$&$39.1\pm1.3$&$42.3\pm1.5$&$44.5\pm1.2$&$45.7\pm1.8$&$44.5\pm1.9$\\
&&LPGNN &$46.8\pm2.0$&$49.7\pm0.7$&$50.7\pm1.4$&$51.2\pm1.4$&$53.3\pm0.9$&$54.5\pm1.2$\\
&&{\em Solitude} &$\mathbf{50.1}\pm1.3$&$\mathbf{52.7}\pm1.0$&$\mathbf{54.0}\pm1.0$&$\mathbf{55.6}\pm0.8$&$\mathbf{56.3}\pm1.0$&$\mathbf{57.0}\pm1.4$\\
\cline{2-9}
&\multirow{3}{*}{GCN}
&Base &$39.7\pm3.0$&$45.3\pm1.8$&$48.3\pm2.9$&$49.6\pm1.3$&$52.5\pm2.1$&$52.8\pm2.5$\\
&&LPGNN &$47.8\pm1.1$&$50.1\pm1.8$&$52.3\pm1.2$&$53.7\pm1.3$&$55.6\pm1.3$&$55.6\pm1.5$\\
&&{\em Solitude} &$\mathbf{49.8}\pm1.4$&$\mathbf{53.4}\pm1.0$&$\mathbf{55.1}\pm0.8$&$\mathbf{56.4}\pm1.5$&$\mathbf{58.2}\pm0.7$&$\mathbf{58.0}\pm2.5$\\
\hline
&&$\epsilon_a$&$7.7$ &$7.9$& $8.0$&$8.3$&$8.5$& $8.7$\\ 
\cline{2-9}
\multirow{6}{*}{LastFM} 
&\multirow{3}{*}{GraphSage}
&Base &$62.6\pm2.4$&$64.4\pm1.8$&$66.5\pm1.9$&$71.2\pm1.9$&$72.0\pm2.0$&$75.5\pm1.4$\\
&&LPGNN &$62.6\pm2.4$&$65.3\pm5.8$&$68.3\pm2.4$&$73.3\pm3.2$&$74.6\pm3.0$&$76.5\pm3.1$\\
&&{\em Solitude} &$\mathbf{65.5}\pm7.1$&$\mathbf{69.4}\pm5.4$&$\mathbf{71.6}\pm2.4$&$\mathbf{77.2}\pm1.7$&$\mathbf{77.2}\pm1.4$&$\mathbf{79.0}\pm1.5$\\
\cline{2-9}
&\multirow{3}{*}{GCN}
&Base &$60.5\pm2.5$&$63.2\pm2.1$&$63.8\pm2.0$&$66.9\pm1.1$&$68.5\pm1.5$&$69.6\pm1.6$\\
&&LPGNN &$\mathbf{62.3}\pm7.2$&$68.2\pm0.8$&$\mathbf{68.3}\pm3.0$&$73.3\pm2.4$&$76.9\pm2.9$&$\mathbf{79.5}\pm0.8$\\
&&{\em Solitude} &$60.9\pm2.6$&$\mathbf{68.9}\pm2.1$&$68.0\pm4.7$&$\mathbf{75.1}\pm1.7$&$\mathbf{77.5}\pm1.9$&$78.9\pm1.3$\\
\hline
\bottomrule
\end{tabular}
\end{sc}
\end{small} 
\end{center}
\vspace{-10pt}
\label{tab:epsx1}
\end{table*}

{\bf Discussions.} In addition to introducing strong baselines for evaluating future improvements to private learning on graphs, our work suggests several open problems and directions for future work:

In this work, the notion of edge-LDP is built upon the randomized mechanism for the neighbor lists represented by a binary vector. This assumption implies that every user and the data curator know how many users and their indexes. We leave the problem of designing new mechanisms that satisfy edge-LDP when the users and the data curator may not know the entire set of the users as future work.  

As illustrated in Sec.~\ref{subsec:ldp}, there are two variants of LDP when applying LDP to graph data: node-LDP and edge-LDP, both of which offer different kinds of privacy protection. In this work, we only consider edge-LDP and LDP for node feature vectors. We believe that private learning on graphs under the notion of node-LDP is a promising direction as well.

Differentially private transfer learning has been studied in prior work and has shown to be a natural candidate for privacy-preserving machine learning in various domains~\cite{abadi2016deep}. As illustrated in the work of {\em Florian et al.}, ~\cite{tramer2021differentially}, the heuristic rule ``better models transfer better'' also holds with differential privacy. Therefore, differentially private graph learning with access to public data from a similar domain may be a feasible solution to improve the privacy-utility guarantees.




\section{Evaluating Privacy-Preserving GNNs}
\label{sec:experiments}

\subsection{Datasets and Experimental Setup}
\label{subsec:settings}

\begin{table}[h!t]
\caption{Statistical Description of Used Datasets.}
\vspace{-10pt}
\begin{center}
\begin{small}\small
\begin{sc}
\setlength{\tabcolsep}{0.1em}
\begin{tabular}{ c| c |c |c| c|c}
\toprule
Dataset &\#Nodes  & \#Edges& \#Classes  & \#Features  & Avg.~Degree\\
\midrule
Cora &$2,708$ & $5,278$ &$7$& $1,433$ & $3.90$\\
Citeseer &$2,110$ & $3,668$ &$6$& $3,703$ & $2.74$\\
LASTFM &$7,083$ & $25,814$ &$10$& $7,842$ & $7.29$\\
\bottomrule
\end{tabular}
\end{sc}
\end{small} 
\end{center}
\vspace{-10pt}

\label{tab:data}
\end{table}

{\bf Datasets.} We conducted experiments over $3$ datasets, falling into two categories: citation networks and social networks. For citation networks, we use Cora~\cite{mccallum2000automating} and CiteSeer~\cite{giles1998citeseer}, both of which are benchmarking datasets for node classification. In these two datasets, nodes represent scientific publications, and edges correspond to the citation links. These two datasets contain bag-of-words feature vectors for each publication; each publication has a class label. For the real-world social networks, we use LastFM~\cite{rozemberczki2020characteristic}, which is a dataset collected from a music streaming service, in which nodes are users from Asian countries, and links represent friendships. Its task is to predict the home country of a user given the artists liked by them. As this dataset was highly imbalanced, for fair comparisons, we limit the classes to the top-10 ones with the most samples, as was done in~\cite{ccs2021}. The detailed statistics of the used datasets are described in Table~\ref{tab:data}.
\begin{figure*}[h!t]
 \centering
 \subfloat[GraphSage on Cora]{\includegraphics[scale = 0.44]{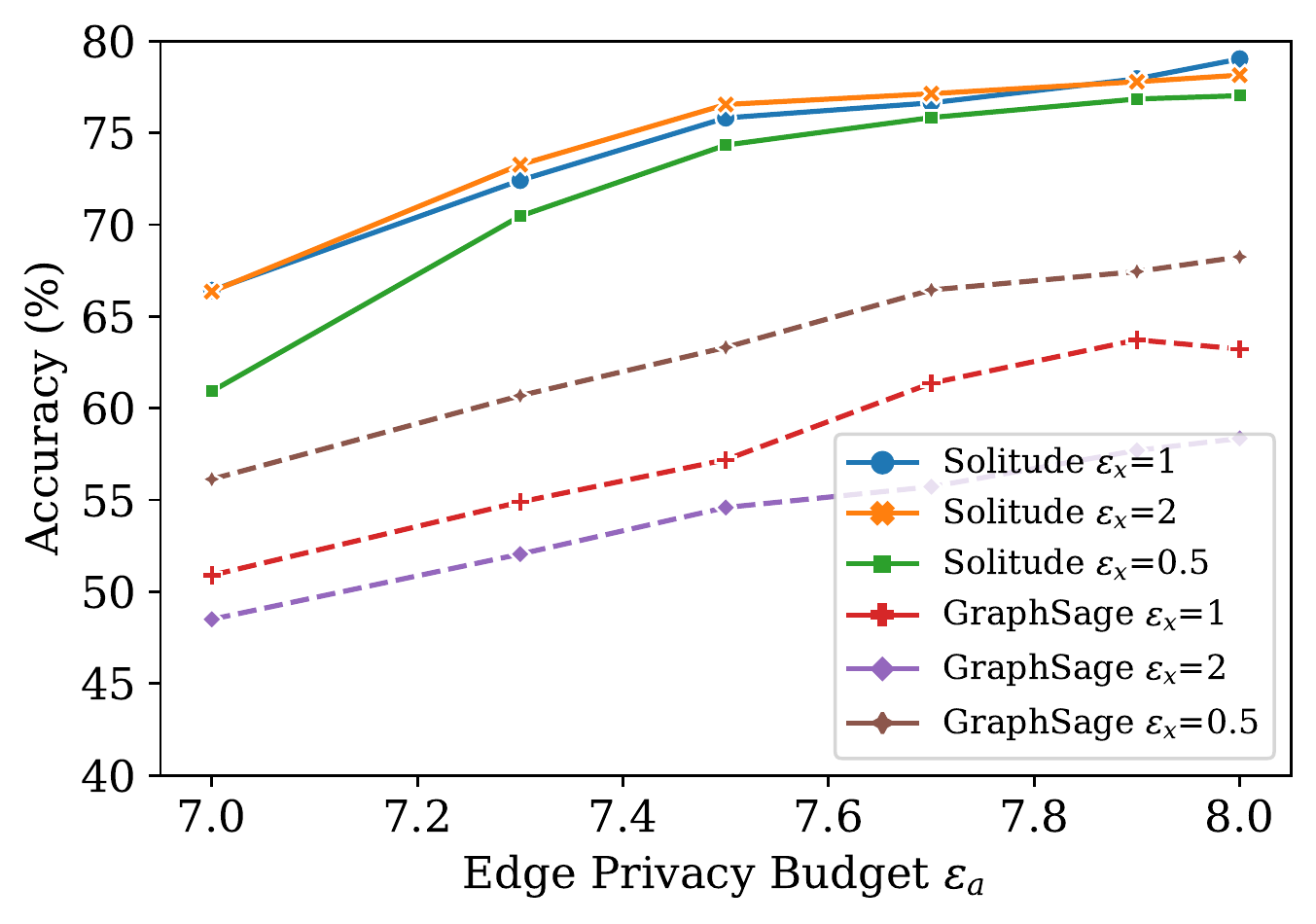}\label{fig:graphsage_cora}}
 \subfloat[GraphSage on Citeseer]{\includegraphics[scale = 0.44]{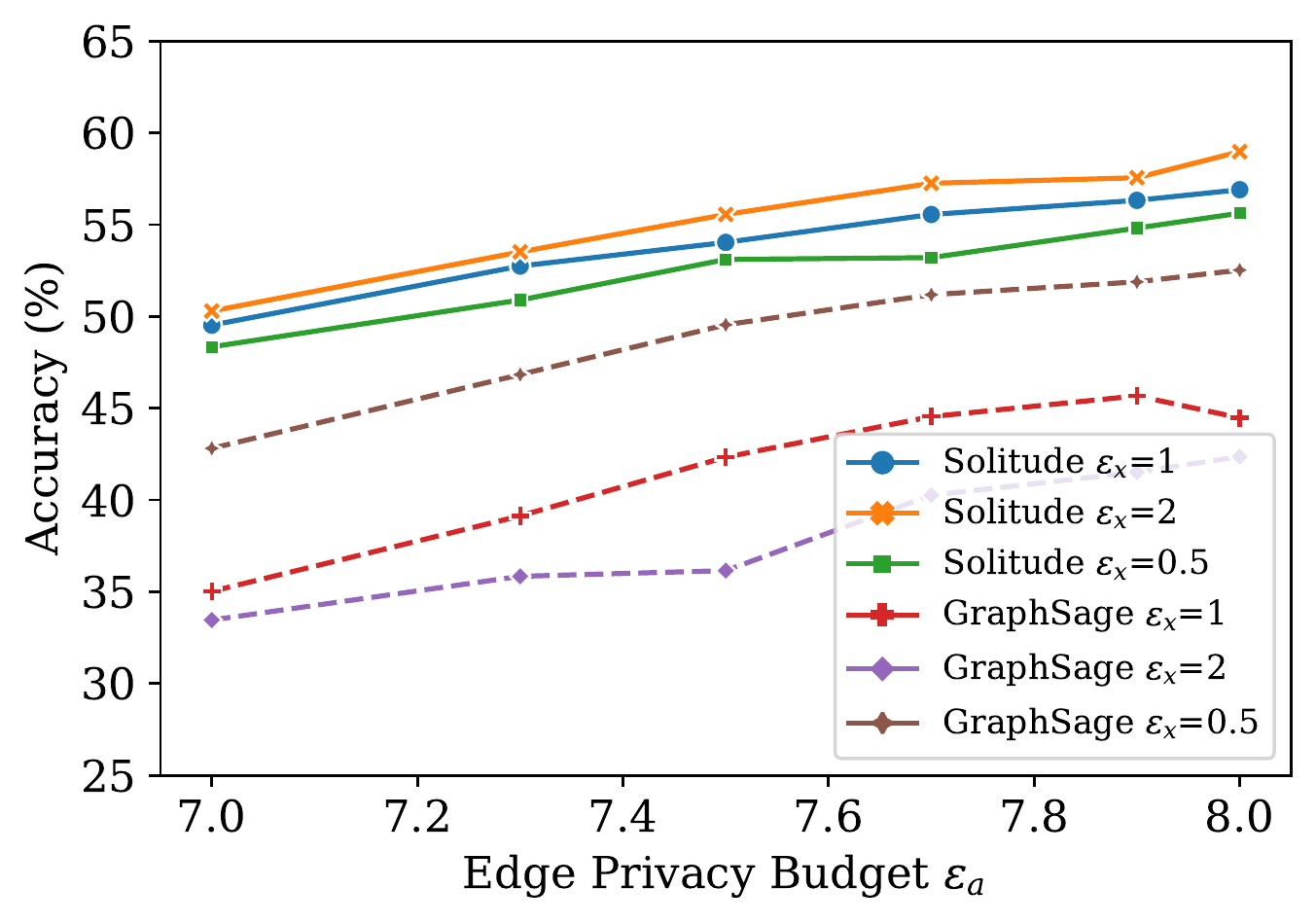}\label{fig:graphsage_citeseer}}
 \subfloat[GraphSage on LastFM]{\includegraphics[scale = 0.44]{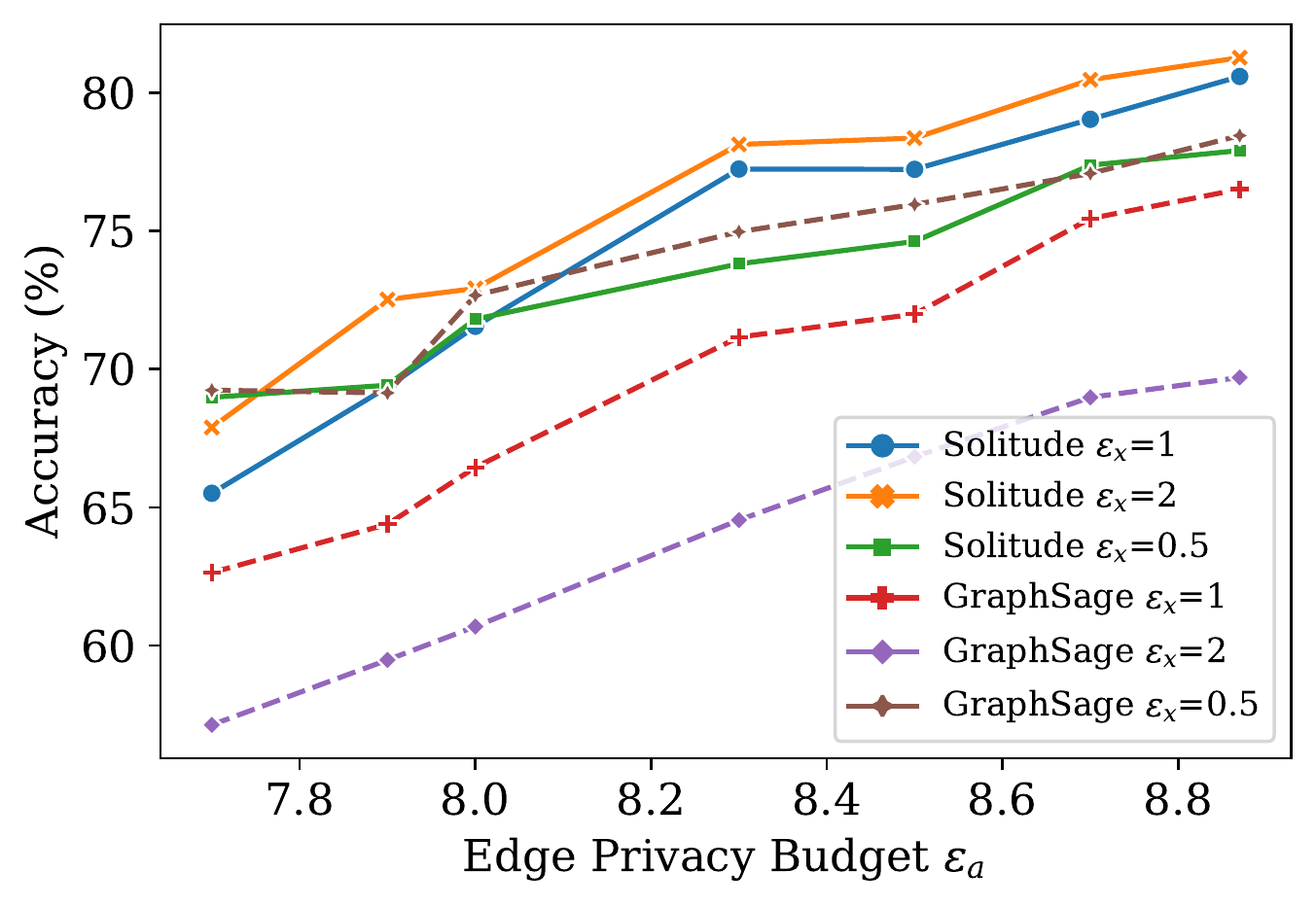}\label{fig:graphsage_lastfm}}\\
 \subfloat[GCN on Cora]{\includegraphics[scale = 0.44]{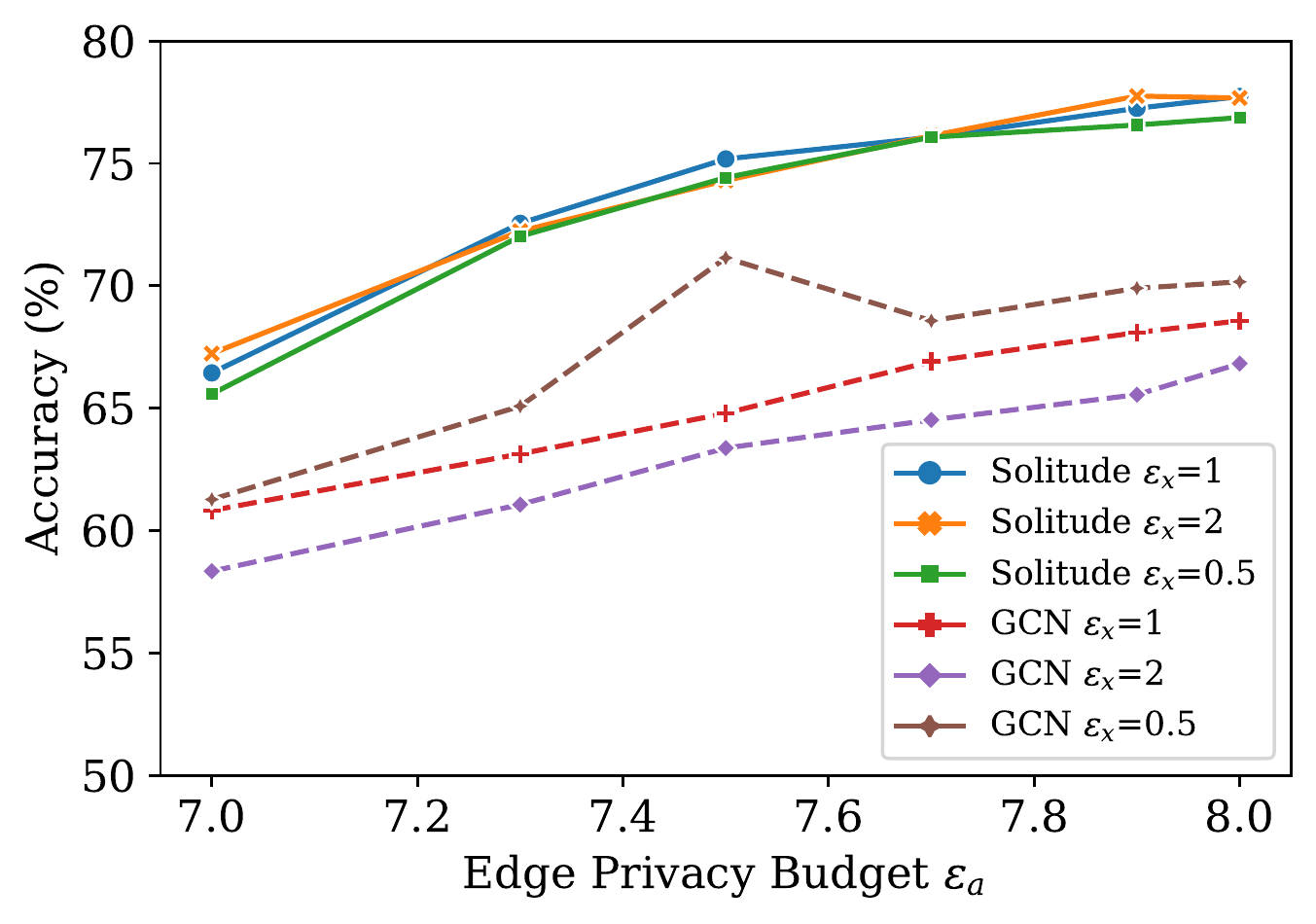}\label{fig:gcn_cora}}
 \subfloat[GCN on Citeseer]{\includegraphics[scale = 0.44]{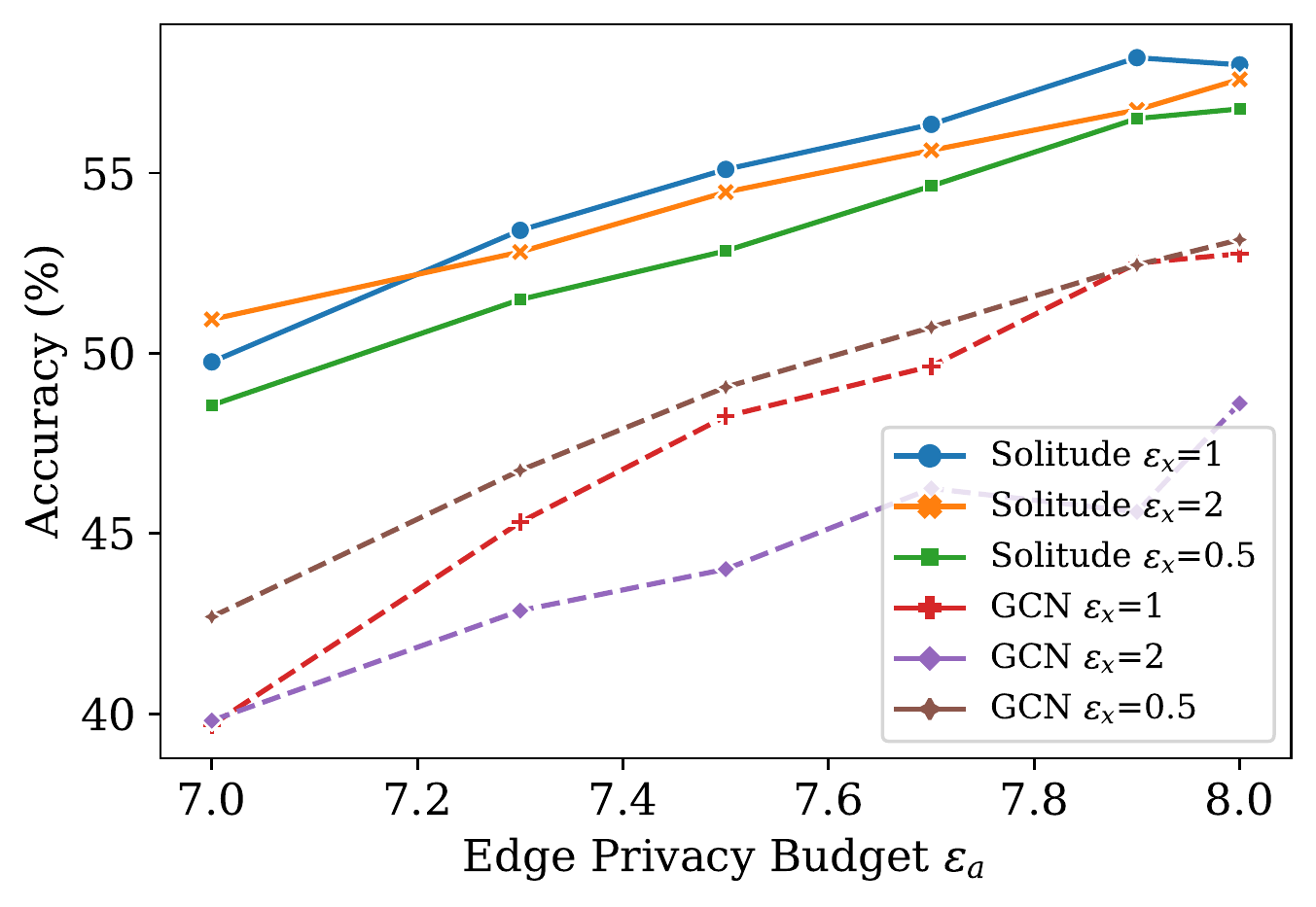}\label{fig:gcn_citeseer}}
 \subfloat[GCN on LastFM]{\includegraphics[scale = 0.44]{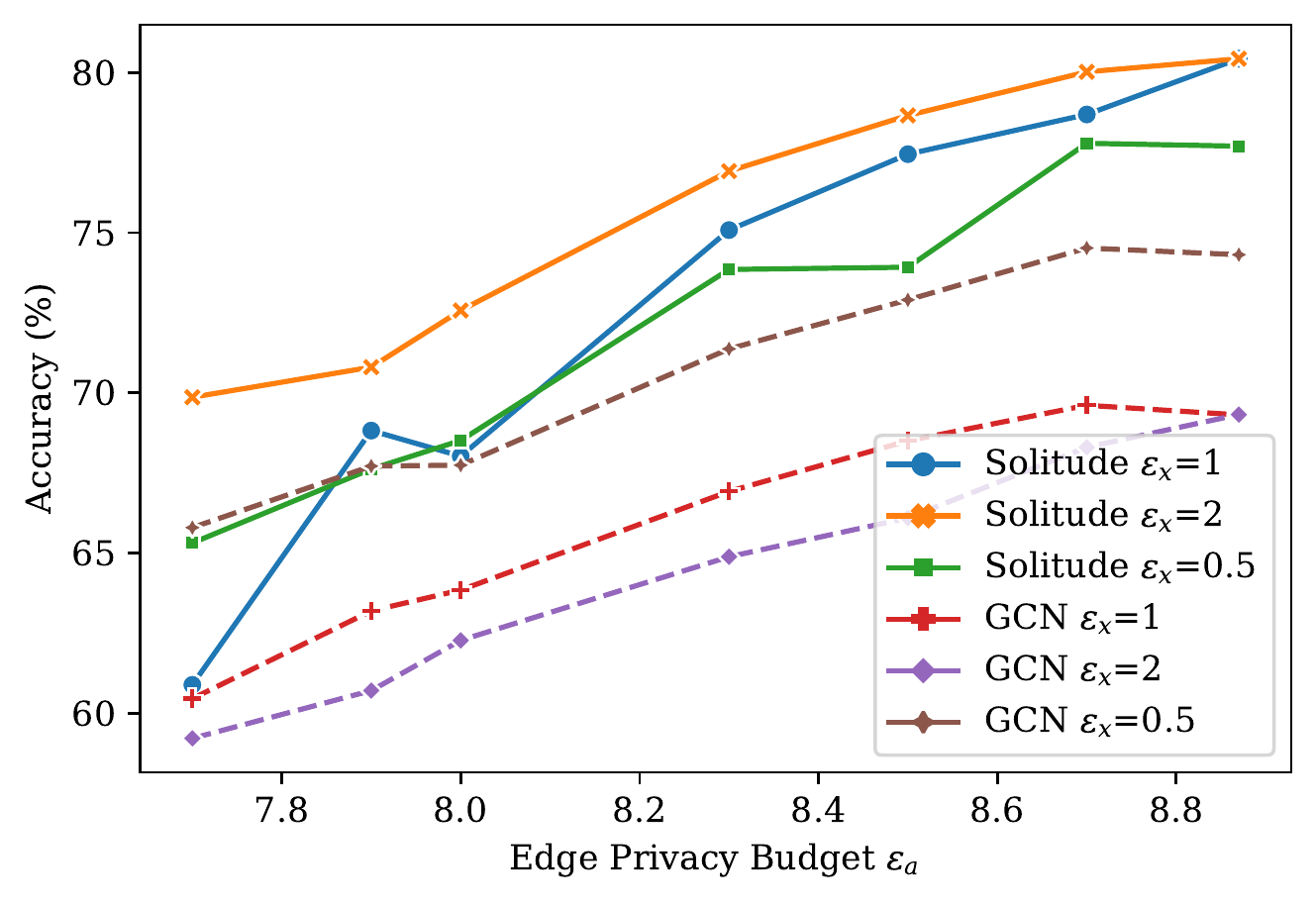}\label{fig:gcn_lastfm}}
  \vspace{-4pt}
 \caption{Classification accuracy with different GNN architectures. The horizontal axes are different values of edge privacy budgets. The vertical axes are the prediction accuracy of the test set. {\em Solitude} obtains significance gains on Cora, Citeseer, and LastFM on various values of $\epsilon_x$ and $\epsilon_a$.}%
 \label{fig:performance}
\end{figure*}

\begin{figure}[h]
   \centering
   \includegraphics[scale = 0.5]{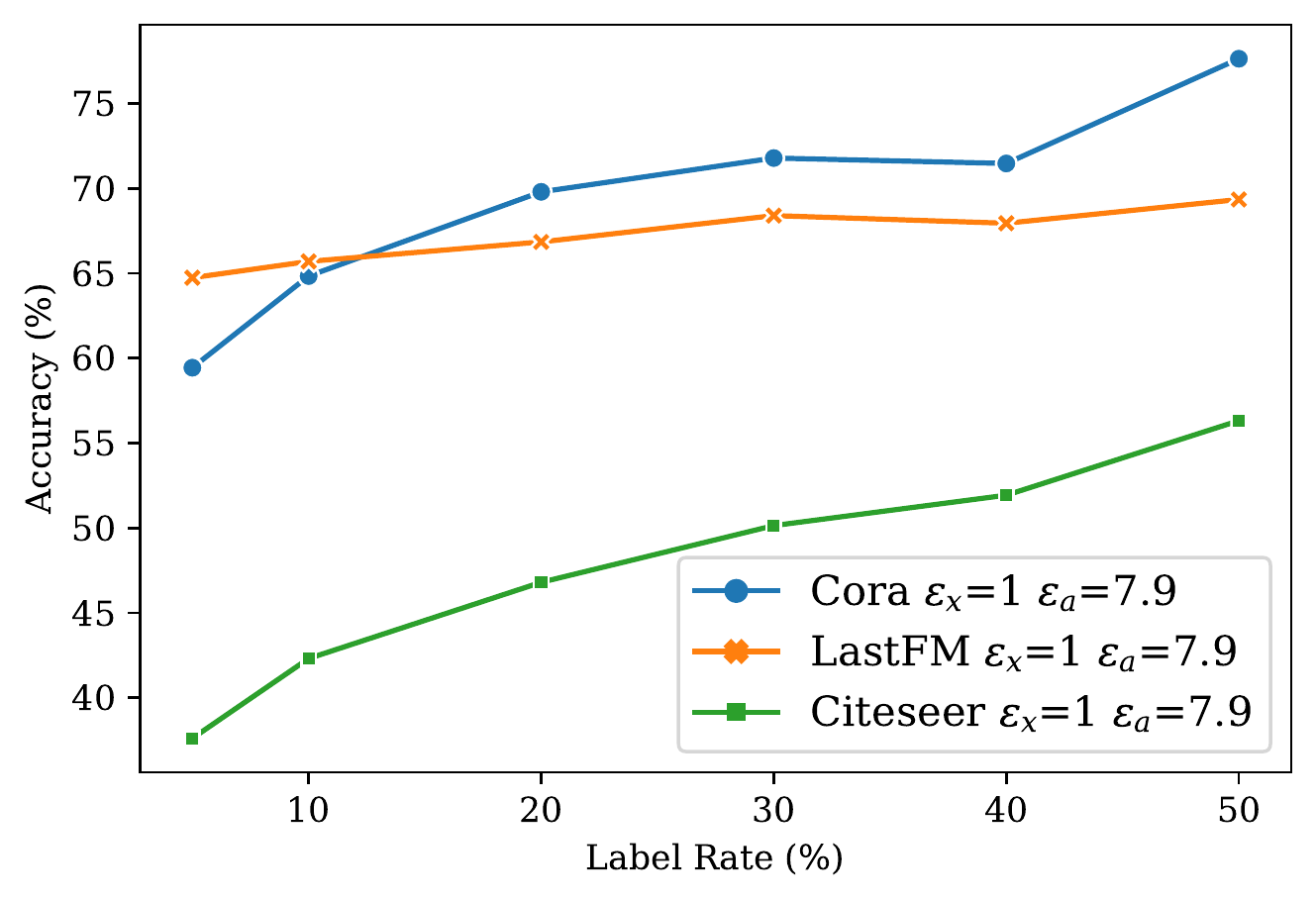}
   \vspace{-10pt}
   \caption{Classification accuracy under various label rates. It shows that improving privacy-utility guarantees needs more labeled samples.}
  \label{fig:label_rate}
\end{figure}

{\bf Baselines.} Our framework is the first effort towards private learning over decentralized network graphs to the best of our knowledge. We compare {\em Solitude} against the {\bf Base} methods including GCN~\cite{kipf2016semi} and GraphSage~\cite{hamilton2017inductive}. In particular, we use the same randomized mechanisms to obfuscate the feature vectors and adjacency lists. The main difference is that the base methods are directly trained over the noisy graphs without any calibration process. For better effectiveness demonstration, we also use an alternative baseline --- {\bf LPGNN}~\cite{ccs2021}. This method assumes that the data curator can access the global topology, and it was proposed to preserve the privacy of the node features. Though the setting of LPGNN and ours are different, to be comparable, we adapt LPGNN into our setting by randomizing the adjacency lists with our proposed mechanism and do not consider the labels of the training samples as the private information of the users. We argue that once the model parameters are achieved, the data curator can obtain the labels for any node by treating them as the model input as long as the model is well trained.
   
{\bf Experimental Setup.} We follow the data preprocessing as reported in~\cite{ccs2021}. Specifically, for all datasets, we randomly split each dataset into three portions: $50\%$ for training, $25\%$ for validation, and $25\%$ for the test. For the datasets with node features, including Cora, CiteSeer, and LastFM, the randomized mechanisms for LDP are applied to the node features and the adjacency lists of all training, validation, and test sets. All the GNN models, including the base methods and the backbone models of LPGNN and {\em Solitude}, consist of two graph convolution layers --- each of which has a hidden dimension of size $16$ --- and a SeLU activation function~\cite{klambauer2017self} followed by dropout.

As for the evaluation metrics, we employ the standard metric --- classification accuracy on the test set (or called prediction accuracy) under various privacy budgets--- to evaluate the generalization capability of the learning model. Unless otherwise stated, all experiments are run $5$ times to ensure statistical significance. Specifically, we report the mean and standard deviation values over $5$ runs.

{\em Parameter settings.} For general hyperparameters, we applied a grid search to find the best choices: the learning rate, dropout rate, and weight decay were tuned among $\{10^{-4}, 10^{-3}, 10^{-2}, 10^{-1}\}$, the feature smoothing steps $l_x$ and label smoothing steps $l_y$ were searched from $\{0,\,2,\,4,\,8\}$, and the coefficients of $\lambda_1$ and $\lambda_2$ were searched in $\{10^{-5},\,10^{-4},\,10^{-3},\,10^{-2}\}$. We used the Xavier initializer~\cite{glorot2010understanding} and the Adam SGD optimizer~\cite{kingma2014adam} for all models. In addition, the maximum epoch was set as $500$. For all $\epsilon_x$ and $\epsilon_a$ pairs, we traversed all the parameters to get its optimal performance in the experimental environment. Without specification, we report the results under the hyperparameters with the best performance overall. More specifically, we use the best learning rate, weight decay, and dropout for every $\epsilon_x$ and $\epsilon_a$ pairs. The selection rationality of the privacy budgets is discussed below.

\subsection{Experimental Results}
\label{subsec:results}

We first investigate how {\em Solitude} performs under varying feature and edge privacy budgets. In particular, the privacy budget for the node features varies within $\{0.5,\, 1, \, 2\}$. The maximum value of $\epsilon_x$ is selected based on the proposition in~\cite{ccs2021}, which indicates that in the high-privacy regime $\epsilon_x\leq2.18$, the multi-bit mechanism perturbs at least one random dimension. Similarly, for randomized response flipping at least one edge of a node, the probability satisfies $p=\frac{1}{1+\e^{\epsilon_a}}\geq\frac{1}{|\mathbf{V}|}$, where $|\mathbf{V}|$ denotes the number of nodes in the graph. Then, we can obtain $\epsilon_a\leq\ln(N-1)$. Therefore, we varies the edge privacy budgets within $\{7.0, 7.3, 7.5, 7.7, 7.9, 8.0\}$ for two citation networks, and $\{7.7, 7.9, 8.0, 8.3, 8.5, 8.7, 8.9\}$ for LastFM. We first fix the privacy budget for the features and compare the performance under various edge privacy budgets. Table~\ref{tab:epsx1} reports the classification accuracy of different methods when $\epsilon_x=1$. 

We can observe that our proposed framework consistently achieves the best performance in most cases. Concretely, {\em Solitude} has an improvement by up to $9.3$ on LastFM as compared to the base methods. As for the two citation networks, our framework achieves significant performance gain ranging from around $7.6\%$ to $18.6\%$ on Cora, and from $5.2\%$ to $15.1\%$ on Citeseer comparing to the base methods (see Table~\ref{tab:epsx1}). The difference in performance gains between LastFM and two citation networks implies that lower degree networks enjoy more benefits from our framework. The reason is that the randomized response has a higher impact on the graphs with more sparseness. Though LPGNN achieves comparable performance gains by comparing with the base methods, it still falls behind our framework in most cases. Note that LPGNN was designed specifically to protect the privacy of node features, while our framework can protect the node features and graph structure information simultaneously. The performance gain achieved by {\em Solitude} upon LPGNN indicates that the denoising component for graph structure is indeed effective in reducing the effects of introduced noise due to the randomized flipping. 

To further examine the effects of introduced noise in the feature vectors, we evaluate the performance of our framework with various privacy budgets for the node features and the adjacency lists. Fig.~\ref{fig:performance} shows the classification accuracy with different GNN architectures across three datasets. We see {\em Solitude} achieves better performance in the lower-privacy regime (with higher privacy budgets). This can be explained by the fact that the predictions become more accurate due to an increase in the privacy budget. Interestingly, we also observe that {\em Solitude} achieves better performance gains with higher privacy budgets for node features when the edge privacy budget is fixed. The maximum performance gain is different across datasets and privacy budgets. This result indicates that our denoising mechanisms effectively mitigate the overfitting issue caused by the introduced noise for privacy protection, improving the model utility.

Essentially, for GNNs to work, the {\em smoothness assumption} has to hold. As the crux of GNNs for node classification is to propagate features and labels throughout the network graph. Inspired by this, we are interested in investigating how the label rates affect the model utility with local differential privacy. In particular, label rate denotes the number of labeled nodes for training divided by the total number of nodes in each dataset. Fig.~\ref{fig:label_rate} depicts the classification accuracy under various label rates. It shows that the prediction accuracy consistently increases with the increase of the label rates over all datasets. We can conclude that improving privacy-utility guarantees needs more labeled nodes for training.

\section{Related Work}
\label{sec:related}

We do not attempt to provide a comprehensive literature review on privacy-preserving graph analytics. Instead, we selectively present the most related methods using differential privacy to preserve the graph data privacy.  

{\bf Privacy-Preserving Analysis of Graph Statistics.} Differential privacy (DP) has emerged as a de-facto standard for privacy guarantees~\cite{raskhodnikova2008can,duchi2013local}. Most existing work focuses on centralized differential privacy~\cite{karwa2011private, wang2013differential}, and they are specialized for analyzing the graph statistics, such as degree distribution estimation~\cite{day2016publishing}, subgraph counts~\cite{karwa2011private}. LDP is a special case of differential privacy in the local model. Each participant obfuscates their data portion locally and then sends the obfuscated data to a data curator (possibly a malicious third-party). Prior works on graph data with LDP mainly focus on the estimations of graph statistics, such as clustering coefficient estimation~\cite{ye2020lf}, heavy hitter estimation~\cite{qin2016heavy}, and frequency estimation~\cite{wang2017locally}. 

Among others, some works are specialized for subgraph counts~\cite{imola2021locally,sun2019analyzing}, such as triangles, $3$-hop paths, and $k$-cliques. Specifically, Sun {\em et al.}~\cite{sun2019analyzing} proposes a multi-phase framework under decentralized differential privacy, which assumes that each user/data holder is aware of not only her connections but also a broader subgraph in her local neighborhood. Subsequently, Imola {\em et al.}~\cite{imola2021locally} propose new algorithms for triangle and $k$-star counts, assuming that each user can only access her connections. LDPGen~\cite{qin2017generating} was proposed to generate synthetic graphs in the setting where a data curator collects subgraphs from the participants under the notion of edge differential privacy.

{\bf Privacy-Preserving Graph Learning with GNNs.} There is a research line that attempts to address privacy in graph learning with federated learning~\cite{meng2021cross,he2021fedgraphnn} and split learning, which is orthogonal to our setting. The closest to ours is LPGNN~\cite{ccs2021}, wherein a set of mechanisms are proposed to protect the privacy of node features. Specifically, LPGNN assumes that a central server holds the global graph topology, and the server is allowed to collect the node features satisfying local differential privacy. It strives to preserve the privacy of node features while maintaining the generalization capability of the learned GNNs. However, the assumption of holding the global topology in the central server hinders its practicality in many real-world applications, where the privacy of the graph topology is of paramount importance. Here we fill this gap and show the effectiveness of our privacy-preserving learning framework based on graph neural networks, with local differential privacy guarantees.



\section{Conclusion}
\label{sec:conclusion}

In this paper, we propose a new privacy-preserving learning framework for decentralized network graphs based on graph neural networks, called {\em Solitude}.  It can simultaneously preserve edge privacy and node feature privacy for every user and seamlessly incorporate with any GNN architectures, such as GCN and GraphSage, with privacy-utility guarantees. The key of {\em Solitude} is a set of new mechanisms that can calibrate the introduced noise in the decentralized graph to mitigate the overfitting problem while learning over noisy graphs. Theoretical analysis and extensive experiments on benchmarks have demonstrated the rationality and effectiveness of our proposed mechanisms.

\bibliographystyle{IEEEtran}
\bibliography{IEEEabrv,main}

\end{document}